\documentclass[5p, twocolumn]{elsarticle}[draft]
\usepackage{amssymb}
\usepackage{amsmath}
\usepackage{diagbox}
\biboptions{numbers,sort&compress}
\usepackage[colorlinks]{hyperref}
\usepackage{caption}
\usepackage[below]{placeins}
\usepackage{color, xcolor}
\usepackage{lineno}
\usepackage{tabularx}
\usepackage{comment}
\usepackage{multirow}
\begin{document}
\begin{frontmatter}
\title{FAIR: Frequency-aware Image Restoration for Industrial Visual Anomaly Detection}
\author[Address1]{Tongkun Liu}
\author[Address1,Address2]{Bing Li}
\author[Address1]{Xiao Du}
\author[Address1]{Bingke Jiang}
\author[Address1]{Leqi Geng}
\author[Address3]{Feiyang Wang}
\author[Address1]{Zhuo Zhao\corref{cor1}}
\cortext[cor1]{Corresponding author at School of Mechanical Engineering, Xi’an Jiaotong University, Xi'an, Shaanxi, China.}
\address[Address1]{State Key Laboratory for Manufacturing System Engineering, Xi’an Jiaotong University, No.99 Yanxiang Road, Yanta District, 710054, Xi’an, Shaanxi, China}

\address[Address2]{International Joint Research Laboratory for Micro/Nano Manufacturing and Measurement Technologies, Xi’an Jiaotong University, No.99 Yanxiang Road, Yanta District, 710054, Xi’an, Shaanxi, China}

\address[Address3]{Anhui Province Key Laboratory of Machine Vision Inspection, Yangtze River Delta HIT Robot Technology Research Institute, Wuhu,24100, China}

\begin{abstract}
 Image reconstruction-based anomaly detection models are widely explored in industrial visual inspection. However, existing models usually suffer from the trade-off between normal reconstruction fidelity and abnormal reconstruction distinguishability, which damages the performance. In this paper, we find that the above trade-off can be better mitigated by leveraging the distinct frequency biases between normal and abnormal reconstruction errors. To this end, we propose \textbf{F}requency-\textbf{a}ware \textbf{I}mage \textbf{R}estoration (FAIR), a novel self-supervised image restoration task that restores images from their high-frequency components. It enables precise reconstruction of normal patterns while mitigating unfavorable generalization to anomalies. Using only a simple vanilla UNet, FAIR achieves state-of-the-art performance with higher efficiency on various defect detection datasets. Code: https://github.com/liutongkun/FAIR.

\end{abstract}


\end{frontmatter}
\begin{figure}[h]
    \centering
		\includegraphics[width=0.98\columnwidth]{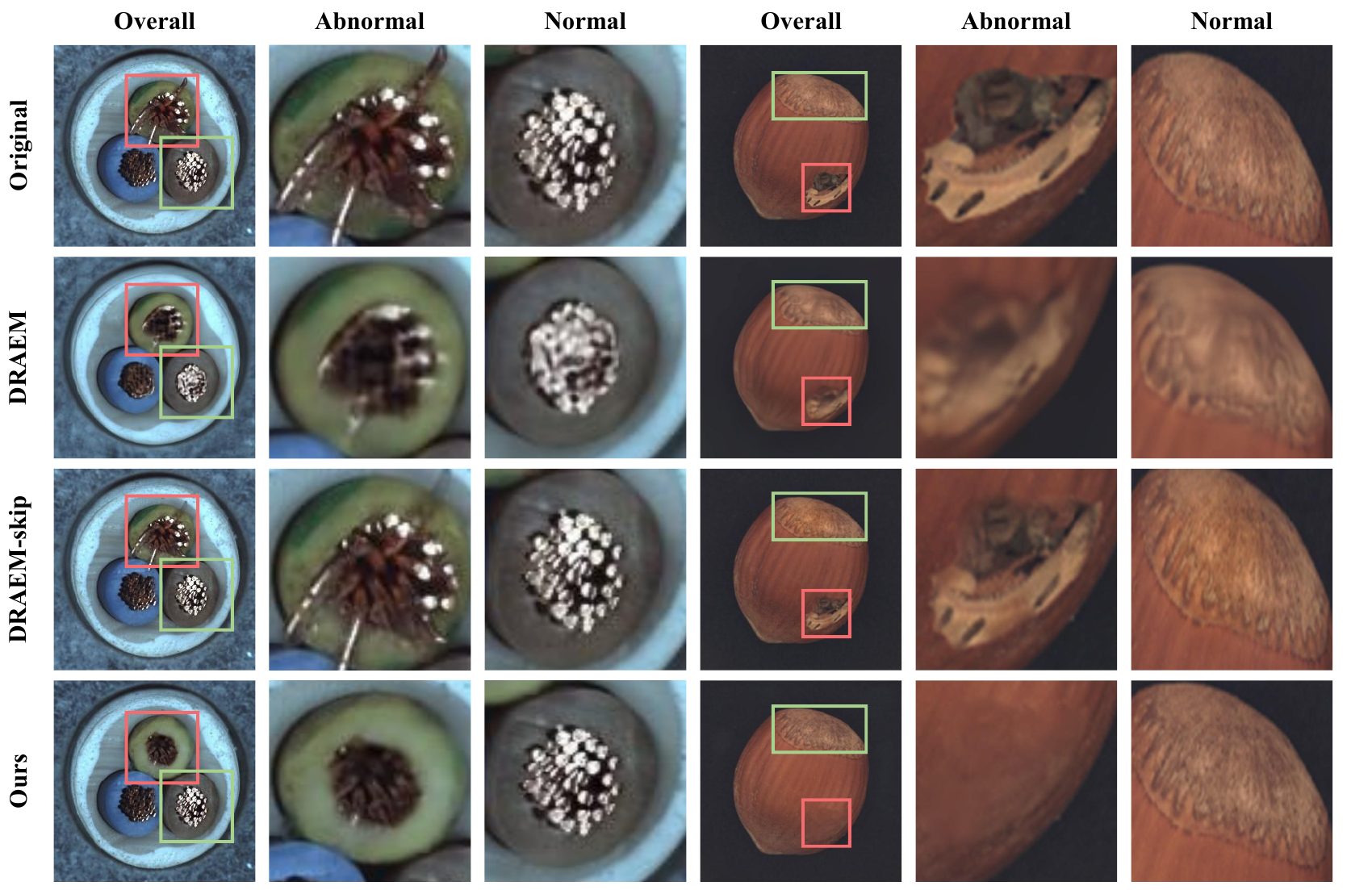}
	\caption{Qualitative comparison of the normal reconstruction fidelity and abnormal reconstruction between DRAEM \cite{zavrtanik2021draem} (without skip connections), DRAEM-skip (with skip connections), and our method.}
	\label{Abstracttradeoff}
\end{figure}   

\begin{figure}[h]
    \centering
		\includegraphics[width=0.98\columnwidth]{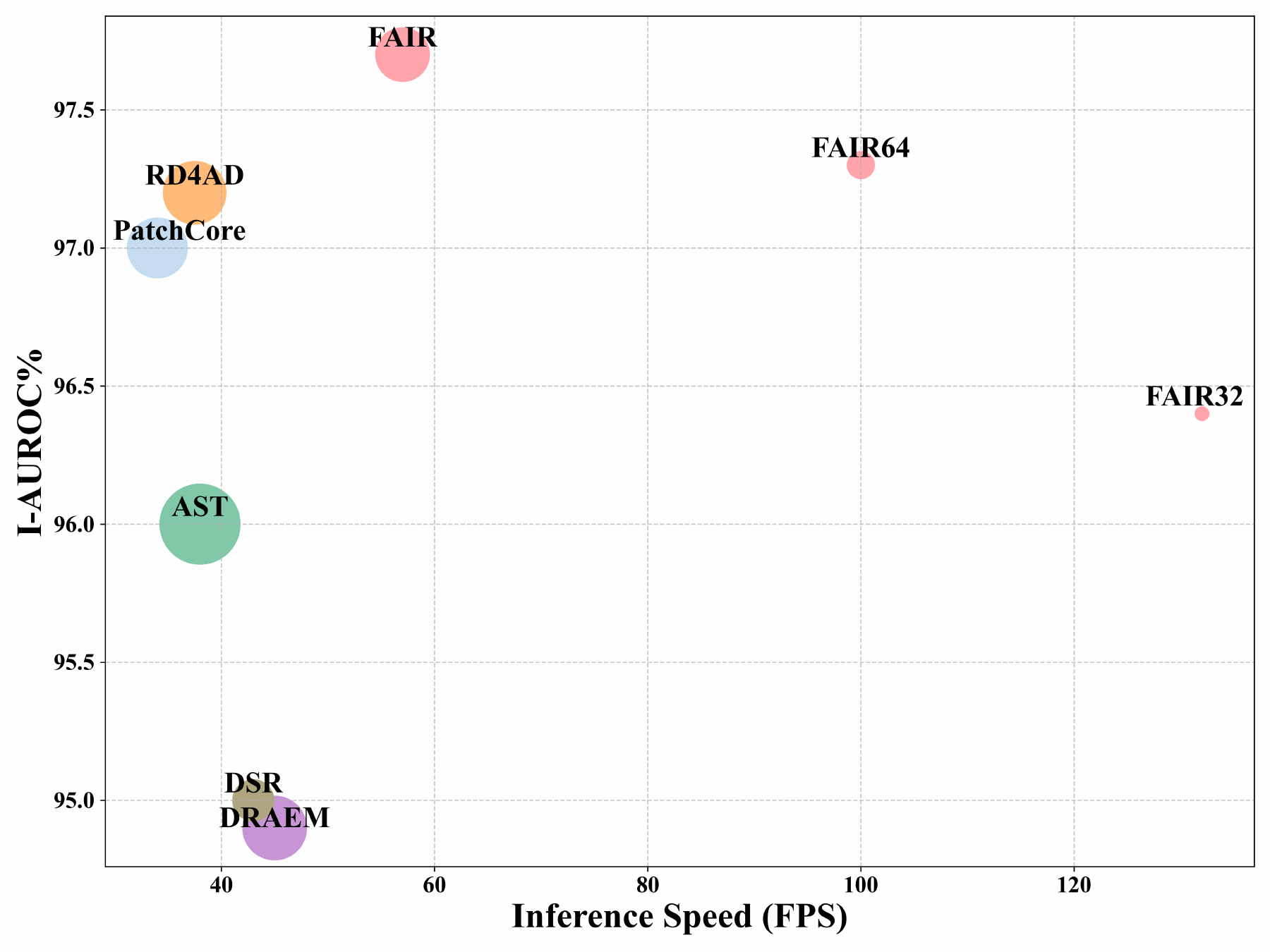}
	\caption{Comparison of the image-level anomaly detection performance and model efficiency with an NVIDIA 3090Ti. The performance is an average value evaluated on the MVTec AD \cite{bergmann2019mvtec} and VisA \cite{zou2022spot} benchmarks. A larger radius of the circle represents more model parameters.}
	\label{AbstractPerformance}
\end{figure}
\section{Introduction}
\label{}
Industrial visual inspection plays an important role in manufacturing quality control. It aims at identifying all types of visible defects that possibly occur on the product. Considering that the defective samples are usually diverse and expensive to collect, supervised models are not well-suited in this field, as these models typically require many defective samples for training and have difficulties in handling those novel types of defects. Therefore, recently, it's preferable to apply visual anomaly detection models \cite{diers2023survey,liu2023deep,horwitz2023back,wang2023multimodal}, which only necessitate normal samples for training and can theoretically detect any types of defects.

Many visual anomaly detection methods are based on image reconstruction \cite{bergmann2018improving,hou2021divide,collin2021improved,zavrtanik2021draem,zavrtanik2022dsr}. These methods assume that the model trained on normal samples can only well reconstruct normal regions but fail in abnormal regions, and therefore the reconstruction errors can be utilized for identifying anomalies. However, in practice, this assumption may not always hold since it's difficult to find a definitive generalization boundary when there is no real defective sample as supervision. In this case, suppressing the model's generalization to anomalies often requires more compact latent representations, which are prone to losing the fine details in normal regions. Conversely, preserving those fine details often requires overparameterized representations or skip connections. Yet, these significantly increase the possibility of identity mapping, resulting in undesired reconstruction of abnormal regions as well.

A possible solution is to convert the conventional image reconstruction task into self-supervised image restoration tasks \cite{ye2020attribute,zavrtanik2021reconstruction,pirnay2022inpainting,huang2019inverse}. These methods first remove some information of the original image and then train a network to restore it. It's expected that the selected removal information can exhibit distinct biases between normal and abnormal data, which help the model establish a more discriminative generalization boundary. However, existing restoration tasks generally fail to find such information. Consequently, the aforementioned trade-off still exists. Specifically, if too much information is removed, the network will encounter challenges in accurately restoring normal regions; instead, if too little information is removed, the preserved abnormal information may be directly propagated to the final restoration results.

To seek such biased information, we start by analyzing the image reconstruction errors in anomaly detection. There are mainly two types of image reconstruction errors: (1). \textbf{Image degradation errors}, which are primarily attributed to the downsampling operation during the encoding phase and the blurring effects induced by Mean Squared Error (MSE) loss. Therefore, these errors typically manifest in \textbf{high-frequency ranges}. (2). \textbf{Anomaly generalization errors}, which arise due to abnormal features not being included in the training set. Given the uncertainty of anomalies, anomalies can manifest in both high-frequency and low-frequency ranges. Consequently, these errors can encompass \textbf{all frequency ranges}. It is worth noting that these two types of errors are often not simply superimposed in the same region, but rather distributed across different regions. Therefore, the reconstruction errors from high-frequency normal regions often overshadow those errors from less prominent anomalies, leading to failure detection cases.

For a model that has been sufficiently trained on normal samples, its normal reconstruction errors typically only contain the image degradation error. This indicates that normal reconstruction errors exhibit a significant bias to high-frequency ranges. However, its abnormal reconstruction errors may come from both types of errors, which are uncertain in the frequency domain. Hence, it can be inferred that, without considering specific types of anomalies, frequency information generally exhibits distinct biases between normal and abnormal data, which can therefore be leveraged to design more suitable image restoration tasks for anomaly detection.

To this end, in this paper, we propose a novel self-supervised restoration task called \textbf{F}requency-\textbf{a}ware \textbf{I}mage \textbf{R}estoration (FAIR) for industrial visual inspection. Our method can achieve more precise normal reconstruction while mitigating unfavorable generalization to anomalies, as shown in Fig. \ref{Abstracttradeoff}. The network architecture is very simple, containing only a vanilla UNet \cite{ronneberger2015u}. Yet, it achieves superior performance along with higher efficiency. We propose to restore the image from its high-frequency component. This design preserves high-frequency information biased towards normal reconstruction errors to enhance normal reconstruction fidelity while removing the low-frequency components, which typically occupy the majority of the image energy, to reduce anomalies' identity mapping. On the other hand, preserving high-frequency information may also lead to identity mapping of high-frequency anomalies. Therefore, during the training phase, the input high-frequency component is corrupted with certain noise to build an additional denoising task. Besides, we conduct a detailed exploration of the impacts of different frequency domain filter designs, which can guide the adjustment of our model in specific domains. Our contributions are summarized as follows: 

(1). We propose a novel image restoration task for industrial visual anomaly detection. It leverages the distinct frequency domain biases between normal and abnormal reconstruction errors, enabling both normal reconstruction fidelity and abnormal reconstruction distinguishability.

(2). We conduct a thorough analysis of the effects of different frequency domain filter designs. This provides a foundation for the adjustment and future optimization of our approach in specific domains.

(3). Our approach achieves state-of-the-art performance on the comprehensive benchmarks MVTec AD \cite{bergmann2019mvtec} and VisA \cite{zou2022spot}. The method only contains a simple UNet, rendering it more efficient and convenient for practical deployment.

\section{Related Work}
\label{20}
Existing industrial visual anomaly detection models can be broadly divided into two groups: feature-based methods and image reconstruction-based methods.

\subsection{Feature-based methods}
\label{2.1}
Feature-based methods aim to find a feature space where normal and abnormal features are fully distinguishable. Since there exists no abnormal sample during training, it's preferable to leverage the ImageNet pre-trained networks \cite{bergmann2020uninformed,roth2022towards,yu2021fastflow,defard2021padim,zheng2022focus,rippel2021modeling,shi2021unsupervised,zhou2022pull,rudolph2023asymmetric} or the models trained from scratch through self-supervised tasks \cite{li2021cutpaste,yi2020patch,sohn2020learning,schluter2022natural}, as feature extractors. For the pre-trained network, several studies \cite{roth2022towards,shi2021unsupervised} find that it's important to select appropriate hierarchy levels of features, because the low-level features lack global awareness, while the very high-level features may be biased to the pre-trained task itself. Many methods use the pre-trained network as a teacher network to detect anomalies by knowledge distillation \cite{bergmann2020uninformed,zhou2022pull,rudolph2023asymmetric}. For the self-supervised based methods, typical tasks include the synthesized anomaly prediction\cite{li2021cutpaste}, the position prediction \cite{yi2020patch}, the geometric transformation prediction \cite{sohn2020learning}, etc. Overall, feature-based methods, especially those leveraging pre-trained models, generally require lower training costs and exhibit higher performance compared to image reconstruction-based methods. However, their detection processes are relatively more implicit, making them difficult to visualize and optimize for specific cases. Meanwhile, those methods relying on pre-trained networks may face challenges related to domain shift.  

\subsection{Image reconstruction-based methods}
\label{2.2}
Image reconstruction-based methods first train image reconstruction models on normal images. Commonly used models include: Autoencoders \cite{bergmann2018improving,collin2021improved,zavrtanik2021reconstruction}, VAEs \cite{venkataramanan2020attention}, GANs \cite{schlegl2019f}, etc. Assuming that the model trained on normal images can not well reconstruct abnormal images, these methods utilize image reconstruction errors to detect anomalies. Compared to feature-based methods, image reconstruction-based methods are generally more interpretable since one can directly observe the intermediate reconstruction results in the explicit image space. However, the key challenge lies in how to balance normal reconstruction fidelity and abnormal reconstruction distinguishability. To solve this issue, some methods \cite{gong2019memorizing,hou2021divide,wu2022self} utilize memory banks that store latent representations of training normal samples to regularize the representations of test samples. While being effective, these memory banks also introduce additional computational overhead due to feature storage and feature addressing. Besides, many works leverage synthesized anomalies to train a denoising model \cite{collin2021improved,zavrtanik2021draem} or along with an additional segmentation model \cite{zavrtanik2021draem}. However, these synthesized texture anomalies may not help address near-in-distribution \cite{zavrtanik2022dsr} anomalies. Therefore, \cite{zavrtanik2022dsr} proposes to leverage quantized features to generate pseudo-anomalies. Yet, the quantized sampling operation significantly augments the training duration.

For the aforementioned reconstruction models, their commonality is that they all use the original clean image during the testing phase. For example, although those denoising models corrupt the image during training, in the testing phase, they still use the original clean image as input. This design preserves all the test image's information. Consequently, the preserved abnormal information may undergo identity mapping into the final output, leading to detrimental results. To this end, some methods propose to convert conventional image reconstruction tasks into image restoration tasks, whose inputs preserve only partial image information in both the training and testing phases. These tasks include geometric transformation restoration tasks\cite{golan2018deep,hendrycks2019using,huang2019inverse}, inpainting tasks \cite{haselmann2018anomaly,zavrtanik2021reconstruction,pirnay2022inpainting} etc., which are summarized into an attribute removal-and-restoration framework by Ye et al. \cite{ye2020attribute}. In \cite{ye2020attribute}, it's observed that the network can learn more discriminative features by restoring the previously removed attributes. However, existing image restoration tasks are generally less effective. For example, the geometric transformation restoration tasks can not be applied to spatially invariant textures. The inpainting tasks suffer from images with random patterns and are typically inefficient since they can only reconstruct part of (the masked part) an image in a forward pass.  

Our approach follows the previous paradigms of image restoration and denoising. Differently, we leverage the frequency domain biases of reconstruction errors to build a novel frequency restoration task. Compared to existing restoration tasks, our task exhibits higher performance and broader applicability including those spatially invariant textures and random patterns. Compared to existing image reconstruction-based methods, we do not require additional modules such as the memory bank, the separate discriminative network, etc., but only use a simple UNet, offering more simplicity and efficiency. Compared to existing feature-based methods, we perform anomaly detection in the interpretable explicit space and do not rely on pre-trained features which may cause the domain-shift issue. Overall, we outperform many state-of-the-art methods, including both feature-based and image reconstruction-based methods on the challenging benchmarks MVTec AD and VisA.

\begin{figure*}[h]
    \centering
		\includegraphics[width=2\columnwidth]{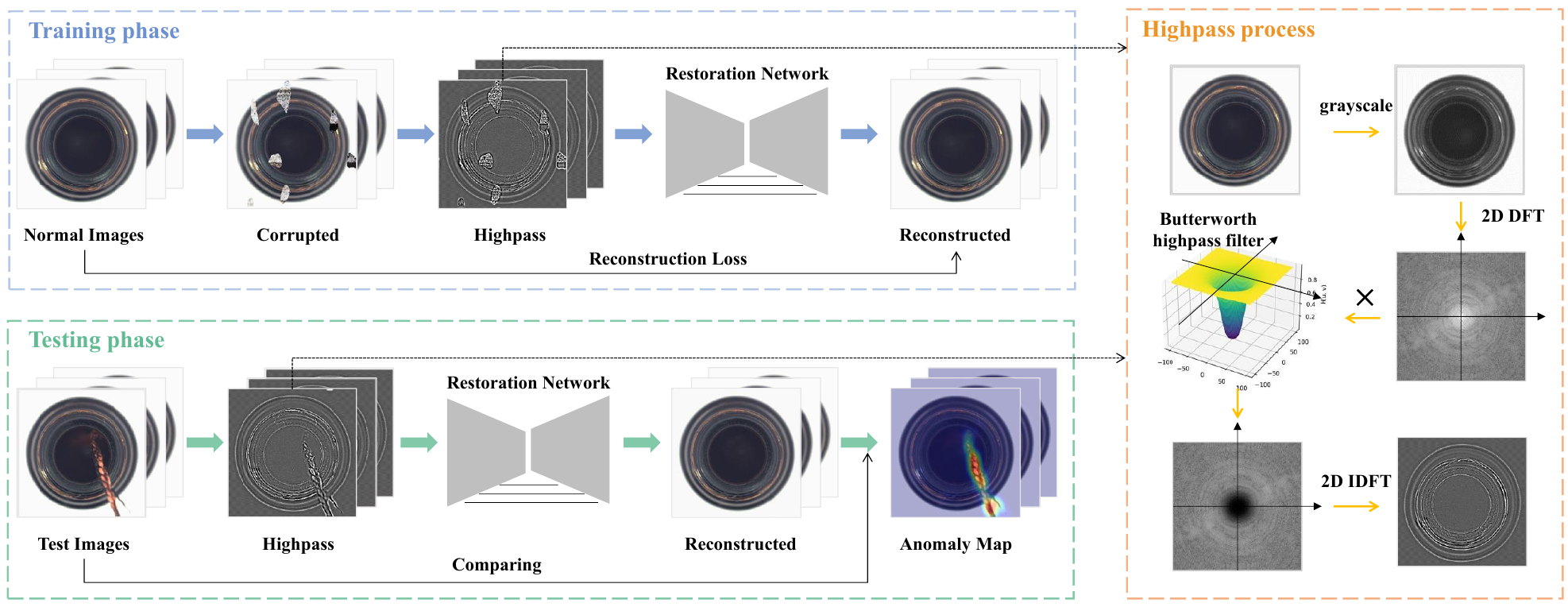}
	\caption{The overview of the proposed method.}
	\label{Overview}
\end{figure*}

\section{Methods}
\label{30}
The overview of the proposed method is shown in Fig. \ref{Overview}. During training, the input normal image is first corrupted with synthesized anomalies and then undergoes a high-pass filter. The network is trained to restore the corrupted high-pass image to the original image. During testing, the input image undergoes the same high-pass filter and is then restored by the network. Finally, an anomaly evaluation function is used to compare the original image with the reconstructed image for anomaly detection.

\subsection{Corrupt the image with synthesized anomalies}
\label{sec3.1}
\begin{figure}[h]
    \centering
		\includegraphics[width=\columnwidth]{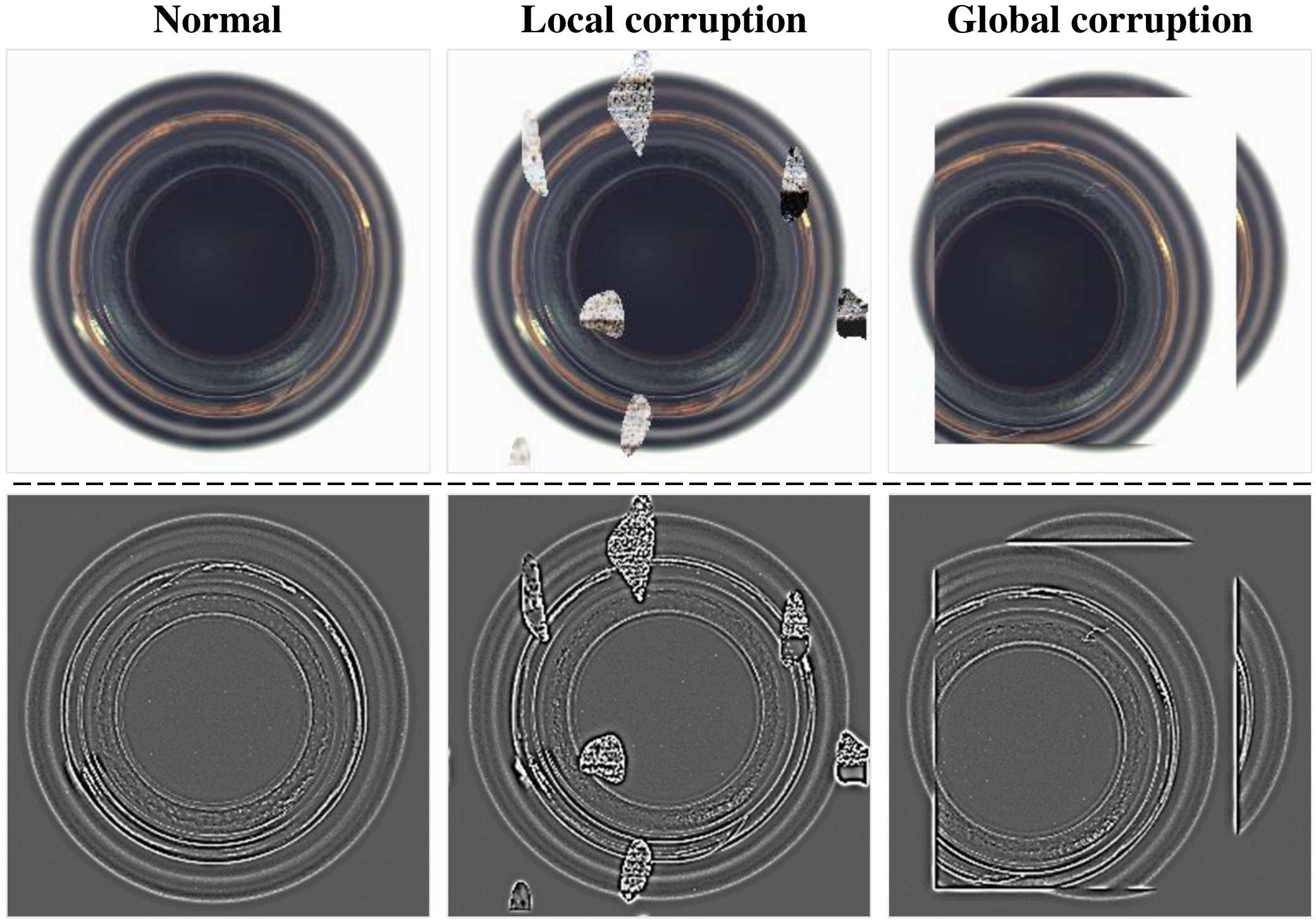}
	\caption{The examples of the local and global synthesized anomalies. The second row shows their highpass results.}
	\label{Synthesizedanomaly}
\end{figure}

Our main principle is preserving high-frequency information and constructing a low-frequency restoration task. However, preserving high-frequency information can lead to the identity mapping of high-frequency anomalies. To mitigate this, we also introduce a conventional denoising task. Given a training normal image $I$, we first corrupt it with the synthesized anomalies. We adopt the strategy proposed in \cite{zavrtanik2021draem} to generate synthesized anomalies whose textures are from the external augmented DTD \cite{cimpoi2014describing} dataset with the shapes of randomly generated Perlin noise, as shown in Fig. \ref{Synthesizedanomaly}, 2nd column. However, due to the significant texture gaps between the external dataset and the target training set, the model can easily tackle such a denoising task by focusing solely on local texture variations. This leads to the model lacking global awareness, making it difficult to detect global anomalies. Consequently, we additionally introduce large CutPaste \cite{li2021cutpaste} augmentations, as shown in Fig \ref{Synthesizedanomaly}, 3rd column. It possesses two characteristics: (1) such synthesized textures are homologous with the training set, thus avoiding the model solely relying on local texture difference for shortcuts; (2) it has an exceptionally large size, which corrupts more than 50\% of the image content, forcing the network to capture global semantics during the denoising process. The corrupted image is denoted as $I^{'}$. The synthesized anomalies are used only in the training phase. 

\subsection{Extract the high-frequency component}
\label{3.2}
Generally, the image's high-frequency information can be extracted either through the spatial domain image derivatives or the frequency domain filtering. In our task, we find both of them to be effective. Here, we first analyze the frequency domain filters and will discuss the spatial domain image derivatives in Sec. \ref{sec5.4}.

\begin{figure}[h]
    \centering
		\includegraphics[width=\columnwidth]{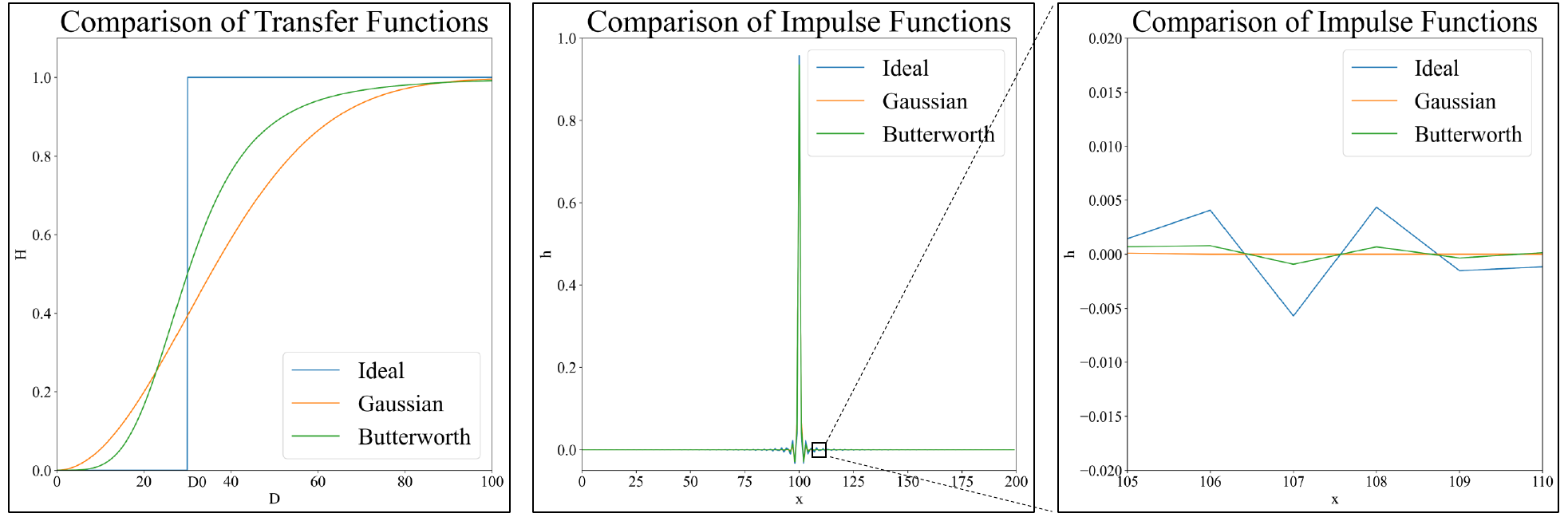}
	\caption{The frequency domain transfer functions and the spatial domain impulse functions of IHPF (Ideal), GHBF (Gaussian), and the second order BHPF (Butterworth). The greater the oscillation of the impulse functions, the more pronounced their ringing artifacts.}
	\label{frequencyplot}
\end{figure}
To perform the frequency domain filtering, the image first undergoes 2D Discrete Fourier transform (DFT), which is:
\begin{equation}
F(u,v)=\sum_{x=0}^{M-1}\sum_{y=0}^{N-1}f(x,y)e^{-i2\pi(\frac{ux}{M}+\frac{vy}{N})}
\end{equation}
where $M, N$ are the image's spatial dimensions. $F(u,v)$ is the image function in the frequency domain and $u, v$ correspond to its coordinates. $f(x,y)$ is the image function in the spatial domain and $x, y$ correspond to its coordinates. We analyze three commonly used filters, including Ideal Highpass Filter (IHPF), Gaussian Highpass Filter (GHPF), and Butterworth Highpass Filter (BHPF). Their respective transfer functions $H_I$, $H_G$, and $H_B$ are:
\begin{equation}
    H_{I}(u,v)=\left\{\begin{matrix}0, \quad    D(u,v)\le D_0
     \\
    1, \quad   D(u,v)>  D_0
    \end{matrix}\right.
\end{equation}

\begin{equation}
        H_{G}(u,v) = 1-e^{-D^{2}(u,v)/2D^{2}_{0}}
\end{equation}

\begin{equation}
      H_{B}(u,v)=\frac{1}{1+[D_0/D(u,v)]^{2n}}
      \label{BHPF}
\end{equation}
where $D(u,v)$ is the Euclidean Distance from any point $(u,v)$ to the origin of the frequency plane, i.e., $D(u,v)=\sqrt{((u-M/2)^2+(v-N/2)^2)}$. $D_0$ is the cutoff frequency corresponding to the transition point between $H(u,v)= 1$ and $H(u,v)=0$. $n$ in eq. \ref{BHPF} refers to the order of BHPF and we originally set it as 2.

\begin{figure}[h]
    \centering
		\includegraphics[width=\columnwidth]{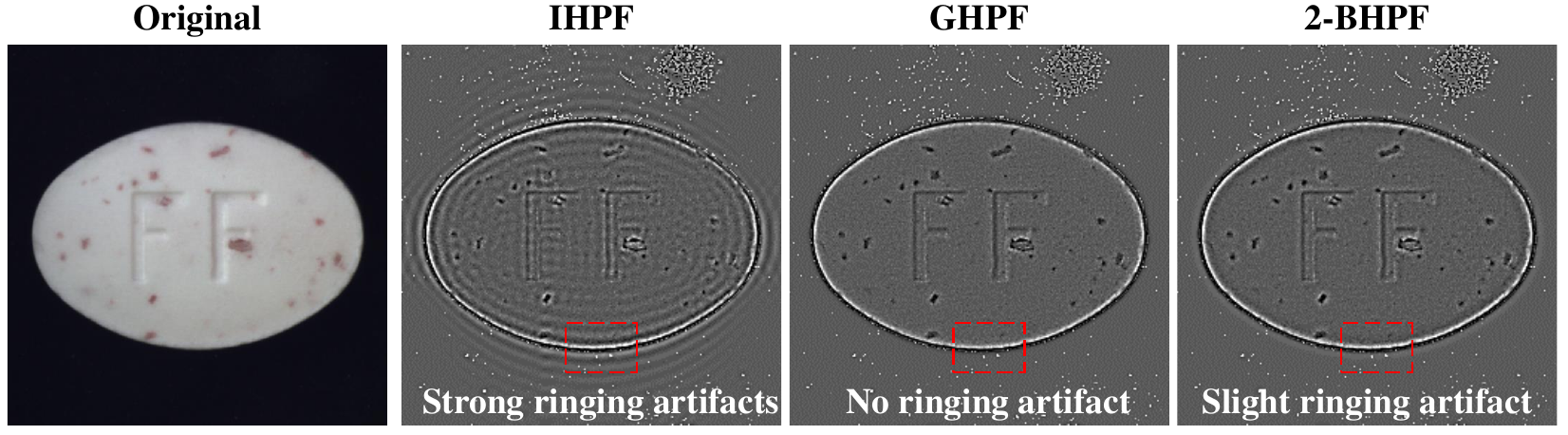}
	\caption{The examples of different spatial ringing artifacts corresponding to different frequency domain filters.}
	\label{Ringingartifact}
\end{figure}

To analyze the impacts of these filters on our restoration task, we plot the frequency domain transfer functions and spatial domain impulse functions of these filters in Fig. \ref{frequencyplot}. Their transfer functions show that IHPF completely suppresses the frequency components below $D_0$, while GHPF just smoothly reduces their amplitudes. BHPF lies in the middle ground. Besides, the impulse functions indicate that these filters generate different levels of ringing artifacts in the spatial domain. Among them, IHPF exhibits the strongest ringing artifacts, while GHPF does not produce any ringing artifacts. BHPF exhibits only weak ringing artifacts. Fig. \ref{Ringingartifact} shows a qualitative comparison. 

For restoration-based anomaly detection models, the above differences correspond to different emphases on normal reconstruction fidelity and abnormal reconstruction distinguishability. Specifically, a hard cutoff of low-frequency information and strong ringing artifacts will make the restoration task very challenging, which is detrimental to the accurate reconstruction of normal regions. Conversely, an over-smoothed cutoff and minimal ringing artifacts will preserve too much of the original image information, which damages the discriminative reconstruction of anomalies since the abnormal information may be directly passed to the final result. To this end, our method originally utilize the second-order BHPF as a compromise. Therefore, the filtering process can be written as: 

\begin{equation}
F(u,v)_h=F(u,v)\cdot H_{B}(u,v)
\end{equation}
where $F(u,v)_h$ represents the highpass image in the frequency domain. Then, we perform IFT on it:
\begin{equation}
I^{'}_{h}=f(x,y)_{h}=\frac{1}{MN}\sum_{u=0}^{M-1}\sum_{v=0}^{N-1}F(u,v)_he^{j2\pi}(\frac{ux}{M}+\frac{vy}{N})
\end{equation}
$I^{'}_{h}$ will serve as the input to the restoration network.

\subsection{Training objectives}
\label{3.2}
We adopt a common loss setting in image reconstruction, that is, a pixel-based $l_2$ loss and a patch-based SSIM loss \cite{bergmann2018improving}. The loss function is: \begin{equation}
L(I,I^{r})= l_2(I,I^{r})+ L_{\rm SSIM} (I,I^{r})
\label{1}
\end{equation}
where $I$ is the original clean normal image and $I^{r}$ is $I^{'}_{h}$'s restoration result.

\subsection{Anomaly Evaluation function}
\label{3.3}
Based on the human perception of defects, we design an anomaly evaluation function containing gradient similarity and color similarity. For the gradient similarity, we apply multi-scale gradient magnitude similarity (MSGMS) proposed in \cite{zavrtanik2021reconstruction}. MSGMS calculates the gradient magnitude similarities (GMS) \cite{xue2013gradient} between the original and reconstructed image at different scales through the image pyramid and takes the average value. For our experiments, we use the bottom two scales of the pyramid. The gradient-based anomaly evaluation function is written as:
\begin{equation}
A_{gradient}=1_{M\times N}-{\rm MSGMS} (I,I^{r})
\label{5}
\end{equation}
To evaluate the color similarity, we propose to first convert the image into CIELAB color space, which is more aligned with the human perception of colors. CIELAB space consists of three components: \textbf{L} for lightness, and \textbf{a} and \textbf{b} for colors. We ignore \textbf{L} to reduce the influence of illumination. Specifically, if the illumination is unstable during the image acquisition process, \textbf{L} in the training set images becomes variable. In this case, under the optimization of MSE loss, the model tends to output an average result, leading to detrimental normal reconstruction errors. In contrast, \textbf{a} and \textbf{b} components are relatively stable to illumination changes. Therefore, for the original image $I$ and the reconstructed image $I_r$, we design our color evaluation function as:
\begin{equation}
A_{color}=(I_a^{r}-I_a)^2+(I_b^{r}-I_b)^2
\label{50}
\end{equation}
Finally, we combine $A_{color}$ and $A_{gradient}$ as our pixel-level anomaly evaluation function $A_{score}$, which is:
\begin{equation}
A_{score}=(kA_{color} + A_{gradient})* f_{\rm mean}
\label{6}
\end{equation}
where $f_{\rm{mean}}$ is a mean smooth filter. $k$ is a scaling factor to make $A_{color}$ and $A_{gradient}$ in the same order of magnitude and adjust the importance of color anomalies. For all the experiments, the kernel size of $f_{\rm{mean}}$ is set as $21$. $k$ is set as $3e^-4$. For the image-level detection score, we choose the maximum value of the anomaly map.

\begin{table*}[h]
\centering
\footnotesize
\caption{Quantitative  comparison of the image-level detection results (AUROC\%) on the MVTec AD dataset.}
\label{MVTec_det}
\scalebox{1}{
\begin{tabular}{ccccccccc}
\hline
Category & Patch-SVDD \cite{yi2020patch} & CutPaste \cite{li2021cutpaste} & NSA \cite{schluter2022natural} & RIAD \cite{zavrtanik2021reconstruction} & Intra \cite{pirnay2022inpainting} & DRAEM \cite{zavrtanik2021draem} & DSR \cite{zavrtanik2022dsr} & FAIR \\ \hline
Carpet & 92.9 & \textbf{100} & 95.6 & 84.2 & 98.8 & 97.0 & \textbf{100} & 99.7 \\
Grid & 94.7 & 99.1 & 99.9 & 99.6 & \textbf{100} & 99.9 & \textbf{100} & 99.7 \\
Leather & 90.9 & \textbf{100} & 99.9 & \textbf{100} & \textbf{100} & \textbf{100} & \textbf{100} & \textbf{100} \\
Tile & 97.8 & 99.8 & \textbf{100} & 93.4 & 98.2 & 99.6 & \textbf{100} & \textbf{100} \\
Wood & 96.5 & 99.8 & 97.5 & 93.0 & 98.0 & 99.1 & 96.3 & \textbf{100} \\ \hline
Avg.tex & 94.5 & 99.7 & 98.6 & 95.1 & 99.0 & 99.1 & 99.3 & \textbf{99.9} \\ \hline
Bottle & 98.6 & \textbf{100} & 97.7 & 99.9 & \textbf{100} & 99.2 & \textbf{100} & \textbf{100} \\
Cable & 90.3 & 96.2 & 94.5 & 81.9 & 84.2 & 91.8 & 93.8 & \textbf{98.1} \\
Capsule & 76.7 & 95.4 & 95.2 & 88.4 & 86.5 & \textbf{98.5} & 98.1 & 97.0 \\
Hazelnut & 92.0 & 99.9 & 94.7 & 83.3 & 95.7 & \textbf{100} & 95.6 & 99.2 \\
Metal Nut & 94.0 & 98.6 & \textbf{98.7} & 88.5 & 96.9 & \textbf{98.7} & 98.5 & 98.0 \\
Pill & 86.1 & 93.3 & 99.2 & 83.8 & 90.2 & 98.9 & 97.5 & \textbf{99.0} \\
Screw & 81.3 & 86.6 & 90.2 & 84.5 & 95.7 & 93.9 & \textbf{96.2} & 91.6 \\
Toothbrush & \textbf{100} & 90.7 & \textbf{100} & \textbf{100} & 99.7 & \textbf{100} & 99.7 & \textbf{100} \\
Transistor & 91.5 & 97.5 & 95.1 & 90.9 & 95.8 & 93.1 & 97.8 & \textbf{98.6} \\
Zipper & 97.9 & 99.9 & 99.8 & 98.1 & 99.4 & \textbf{100} & \textbf{100} & 98.5 \\ \hline
Avg.obj & 90.8 & 95.8 & 96.5 & 89.9 & 94.4 & 97.4 & 97.7 & \textbf{98.0} \\ \hline
Avg.all & 92.1 & 97.1 & 97.2 & 91.7 & 95.9 & 98.0 & 98.2 & \textbf{98.6} \\ \hline
\end{tabular}}
\end{table*}

\begin{table*}[htb]
\centering
\footnotesize
\caption{Quantitative comparison of the image-level detection results (AUROC\%) on the MVTec AD dataset.}
\label{MVTec_pixel}
\setlength{\tabcolsep}{5mm}{
\scalebox{1}{
\begin{tabular}{ccccccc}
\hline
Category & PatchCore \cite{roth2022towards} & RD4AD \cite{deng2022anomaly} & AST \cite{rudolph2023asymmetric} & DRAEM \cite{zavrtanik2021draem}& DSR \cite{zavrtanik2022dsr}& FAIR \\ \hline
Candle & \textbf{98.6} & 94.3 & 96.7 & 92.7 & 87.8 & 96.0 \\
Capsules & 80.0 & 90.8 & 83.9 & 90.2 & \textbf{96.4} & 93.9 \\
Cashew & \textbf{97.6} & 97.4 & 97.5 & 85.5 & 92.4 & 96.5 \\
Chewing gum & 99.6 & 98.4 & \textbf{99.8} & 95.2 & 98.4 & 96.2 \\
Fryum & 96.4 & 96.2 & 87.9 & 88.6 & 96.4 & \textbf{96.8} \\
Macaroni1 & 97.3 & \textbf{98.6} & 92.2 & 94.2 & 92.5 & 97.9 \\
Macaroni2 & 78.0 & 89.5 & \textbf{94.4} & 86.6 & 82.7 & 91.7 \\
PCB1 & 98.4 & 97.1 & \textbf{99.4} & 75.9 & 90.7 & 97.7 \\
PCB2 & 97.2 & 97.0 & 95.9 & \textbf{98.9} & 96.4 & 97.2 \\
PCB3 & 98.0 & 96.4 & 96.4 & 94.4 & 97.3 & \textbf{99.3} \\
PCB4 & \textbf{99.7} & 99.8 & 98.0 & 98.6 & 97.9 & 98.7 \\
Pipe fryum & \textbf{99.9} & 94.6 & 99.7 & 97.6 & 96.9 & 98.8 \\ \hline
Avg.all & 95.0 & 95.8 & 92.7 & 91.7 & 91.8 & \textbf{96.7} \\ \hline
\end{tabular}}}
\end{table*}

\begin{table*}[h]
\centering
\footnotesize
\caption{Quantitative comparison of the pixel-level localization results (AUROC\%, AUPRO\%) on the MVTec AD dataset.}
\label{Visa_image}
\begin{tabular}{ccccccccc}
\hline
Category & Patch-SVDD \cite{yi2020patch} & CutPaste \cite{li2021cutpaste} & NSA \cite{schluter2022natural} & RIAD \cite{zavrtanik2021reconstruction} & Intra \cite{pirnay2022inpainting} & DRAEM \cite{zavrtanik2021draem} & DSR \cite{zavrtanik2022dsr} & FAIR \\ \hline
Carpet & (92.6,-) & (98.3,-) & (95.5,85.0) & (96.3,-) & (99.2,-) & (96.2,92.9) & (95.9,93.2) & (\textbf{99.6,98.4}) \\
Grid & (96.2,-) & (97.5,-) & (99.2,96.8) & (98.8,-) & (99.4,-) & (\textbf{99.6},98.4) & (\textbf{99.6},\textbf{98.7}) & (99.4,97.7) \\
Leather & (97.4,-) & (99.5,-) & (99.5,98.7) & (99.4,-) & (99.5,-) & (98.9,97.8) & (99.4,97.6) & (\textbf{99.6},\textbf{98.9}) \\
Tile & (91.4,-) & (90.5,-) & (99.3,95.3) & (89.1,-) & (94.4,-) & (\textbf{99.5},\textbf{98.5}) & (98.8,96.7) & (98.4,95.4) \\
Wood & (90.8,-) & (95.5,-) & (90.7,85.3) & (85.8,-) & (90.5,-) & (97.2,93.5) & (91.8,86.5) & (\textbf{97.3},\textbf{94.2}) \\ \hline
Avg.tex & (93.7,-) & (96.3,-) & (96.8,92.2) & (93.9,-) & (96.6,-) & (98.3,96.2) & (97.1,94.5) & (\textbf{98.9},\textbf{96.9}) \\ \hline
Bottle & (98.1,-) & (97.6,-) & (98.3,92.9) & (98.4,-) & (97.1,-) & (\textbf{99.3},\textbf{97.0}) & (98.7,95.2) & (98.3,94.1) \\
Cable & (96.8,-) & (90.0,-) & (96.0,89.9) & (84.2,-) & (93.2,-) & (95.4,75.6) & (97.4,86.0) & (\textbf{98.5},\textbf{92.3}) \\
Capsule & (95.8,-) & (97.4,-) & (97.6,\textbf{91.4}) & (92.8,-) & (\textbf{97.7},-) & (94.0,91.0) & (91.0,86.0) & (93.9,83.6) \\
Hazelnut & (97.5,-) & (97.3,-) & (97.6,93.6) & (96.1,-) & (98.3,-) & (\textbf{99.5},\textbf{98.6}) & (99.0,92.7) & (99.4,96.0) \\
Metal Nut & (98.0,-) & (93.1,-) & (98.4,\textbf{94.6}) & (92.5,-) & (93.3,-) & (\textbf{98.7},94.0) & (93.0,90.2) & (98.1,88.7) \\
Pill & (95.1,-) & (95.7,-) & (\textbf{98.5},\textbf{96.0}) & (95.7,-) & (98.3,-) & (97.6,88.2) & (93.9,94.3) & (98.4,95.0) \\
Screw & (95.7,-) & (96.7,-) & (96.5,90.1) & (98.8,-) & (99.5,-) & (\textbf{99.7},\textbf{98.2}) & (98.0,91.2) & (98.8,93.2) \\
Toothbrush & (98.1,-) & (98.1,-) & (94.9,90.7) & (98.9,-) & (99.0,-) & (98.1,90.3) & (\textbf{99.4},\textbf{95.3}) & (99.2,94.6) \\
Transistor & (\textbf{97.0},-) & (93.0,-) & (88.0,75.3) & (87.7,-) & (96.1,-) & (90.0,81.4) & (78.2,78.1) & (95.4,\textbf{90.0}) \\
Zipper & (95.1,-) & (99.3,-) & (94.2,89.2) & (97.8,-) & (99.2,-) & (98.6,96.2) & (98.3,94.1) & (\textbf{99.4},\textbf{98.1}) \\ \hline
Avg.obj & (96.7,-) & (95.8,-) & (96.8,90.4) & (94.4,-) & (97.2,-) & (97.1,91.1) & (94.7,90.3) & (\textbf{97.9},\textbf{92.6}) \\ \hline
Avg.all & (95.7,-) & (96.0,-) & (96.3,91.0) & (94.2,-) & (97.0,-) & (97.5,92.8) & (95.5,91.7) & (\textbf{98.2},\textbf{94.0}) \\ \hline
\end{tabular}
\end{table*}

\begin{table*}[h]
\footnotesize
\centering
\caption{Quantitative comparison of the pixel-level localization results (AUROC\%, AUPRO\%) on the VisA dataset.}
\label{Visa_pixel}
\setlength{\tabcolsep}{5mm}{
\scalebox{1}{
\begin{tabular}{ccccccc}
\hline
Category & PatchCore \cite{roth2022towards} & RD4AD \cite{deng2022anomaly} & AST \cite{rudolph2023asymmetric} & DRAEM \cite{zavrtanik2021draem}& DSR \cite{zavrtanik2022dsr}& FAIR \\ \hline
Candle & (\textbf{99.4},\textbf{94.6}) & (98.7,93.1) & (97.1,90.8) & (93.6,92.7) & (84.8,79.7) & (98.0,92.5) \\
Capsules & (\textbf{99.5},87.7) & (99.4,96.0) & (96.7,72.7) & (99.1,85.4) & (96.8,74.5) & (93.9,\textbf{96.2}) \\
Cashew & (\textbf{98.9},\textbf{95.0}) & (94.1,88.3) & (97.5,68.1) & (81.6,67.6) & (97.6,61.5) & (98.8,93.4) \\
Chewing gum & (\textbf{99.1},\textbf{84.5}) & (97.4,82.3) & (98.5,87.1) & (97.5,58.0) & (93.8,58.2) & (98.3,78.8) \\
Fryum & (93.7,85.5) & (96.7,\textbf{91.4}) & (85.0,57.2) & (73.8,80.4) & (82.9,65.5) & (\textbf{96.8},88.9) \\
Macaroni1 & (\textbf{99.8},96.4) & (99.6,95.3) & (96.8,85.3) & (99.4,86.3) & (87.3,57.7) & (99.7,\textbf{97.8}) \\
Macaroni2 & (\textbf{99.1},94.8) & (99.2,96.9) & (90.9,83.0) & (99.0,96.3) & (83.4,52.2) & (91.7,\textbf{98.3)} \\
PCB1 & (\textbf{99.9},94.3) & (99.7,\textbf{96.0}) & (99.4,90.0) & (97.9,61.1) & (90.5,61.3) & (99.7,91.0) \\
PCB2 & (\textbf{99.0},89.9) & (98.6,92.3) & (95.9,71.8) & (97.3,76.2) & (96.5,84.9) & (98.9,\textbf{90.9}) \\
PCB3 & (\textbf{99.3},90.8) & (99.2,\textbf{94.8}) & (96.4,71.3) & (96.0,83.5) & (94.8,79.5) & (\textbf{99.3},92.9) \\
PCB4 & (\textbf{98.4},89.5) & (\textbf{98.4},\textbf{90.3}) & (90.5,64.8) & (94.9,73.4) & (93.5,62.1) & (97.8,80.8) \\
Pipe fryum & (99.1,96.0) & (98.7,\textbf{96.4}) & (98.3,90.5) & (87.1,74.6) & (97.5,80.5) & (\textbf{99.4},95.0) \\ \hline
Avg.all & (\textbf{98.8},91.6) & (98.3,\textbf{92.8}) & (95.3,77.7) & (93.1,78.0) & (91.6,68.1) & (\textbf{98.8},91.4) \\ \hline
\end{tabular}}}
\end{table*}

\section{Experiments}
\subsection{Datasets}
\label{4.1}
We conduct the experiments on the famous MVTec AD \cite{bergmann2019mvtec} and VisA datasets \cite{zou2022spot}. They are both comprehensive benchmarks involving various industrial products and anomaly types. Specifically, MVTec AD covers 15 different industrial products including 5 texture categories and 10 object categories. VisA covers 12 different object categories, which can be divided into three types, including objects with complex structures, objects with multiple instances, and objects with unaligned poses. For both of these benchmarks, their training sets only contain normal samples while the testing sets contain both normal and abnormal samples.

\subsection{Evaluation metrics}
We use the area under the curve (AUC) of the receiver operating characteristics (ROC) to evaluate the performances of image-level anomaly detection and pixel-level anomaly localization. Considering many anomalies only occupy a small fraction of pixels in the image, the AUROC may not well reflect the actual anomaly localization performance. Therefore, we also use the area under the per-region-overlap curve (AUC-PRO) proposed by \cite{bergmann2020uninformed}. Specifically, when using the AUC-PRO, we follow the standard setting where the value is integrated from 0 to 0.3 across FPRs (average per-pixel false positive rate). 

\subsection{Implementation Details}
Our work is implemented in Pytorch with an NVIDIA GeForce GTX 3090Ti and I7-12700. We resize all the images of the MVTec AD and VisA to $256\times 256$ for both training and testing without using any center crop. For the cutoff frequency $D_0$, we originally set $D_0=30$. The restoration network is based on the vanilla UNet with skip connections and we set its base channel with $c$ as $c=128$. On the MVTec AD dataset, we train the model from scratch for 800 epochs with a batch size of 8. On the VisA dataset, as there are more training images in each category, we reduce the epoch number to 400 to keep a similar total iterations. All the experiments use the Adam optimizer with an initial learning rate of $1e^-4$, which is further multiplied by 0.2 after both the 80th and 90th percentile of the epochs. We run the experiments for five times and take the average results.

We choose feature-based methods including Patch-SVDD \cite{yi2020patch}, CutPaste \cite{li2021cutpaste}, NSA \cite{schluter2022natural}, PatchCore \cite{roth2022towards}, RD4AD \cite{deng2022anomaly}, and AST \cite{rudolph2023asymmetric} and image reconstruction-based methods including RIAD \cite{zavrtanik2021reconstruction}, Intra \cite{pirnay2022inpainting}, DRAEM \cite{zavrtanik2021draem}, and DSR \cite{zavrtanik2022dsr} as our baselines. On the MVTec AD dataset, we directly report their original announced performance. Meanwhile, we employ their officially pre-trained weights to assess the metrics that are not stated in their original paper. On the VisA dataset, we re-implement PatchCore, RD4AD, AST, and DRAEM with their official codes and original settings but disable the center crop operation since it will ignore the anomalies at the image boundary. In particular, for Patchcore, we choose the WideResNet-50 \cite{he2016deep} as its backbone. Besides, for DSR, considering its huge training overhead, we use the re-implementation results on the VisA dataset produced by \cite{batzner2023efficientad}.

\begin{figure*}[h]
    \centering
		\includegraphics[width=2\columnwidth]{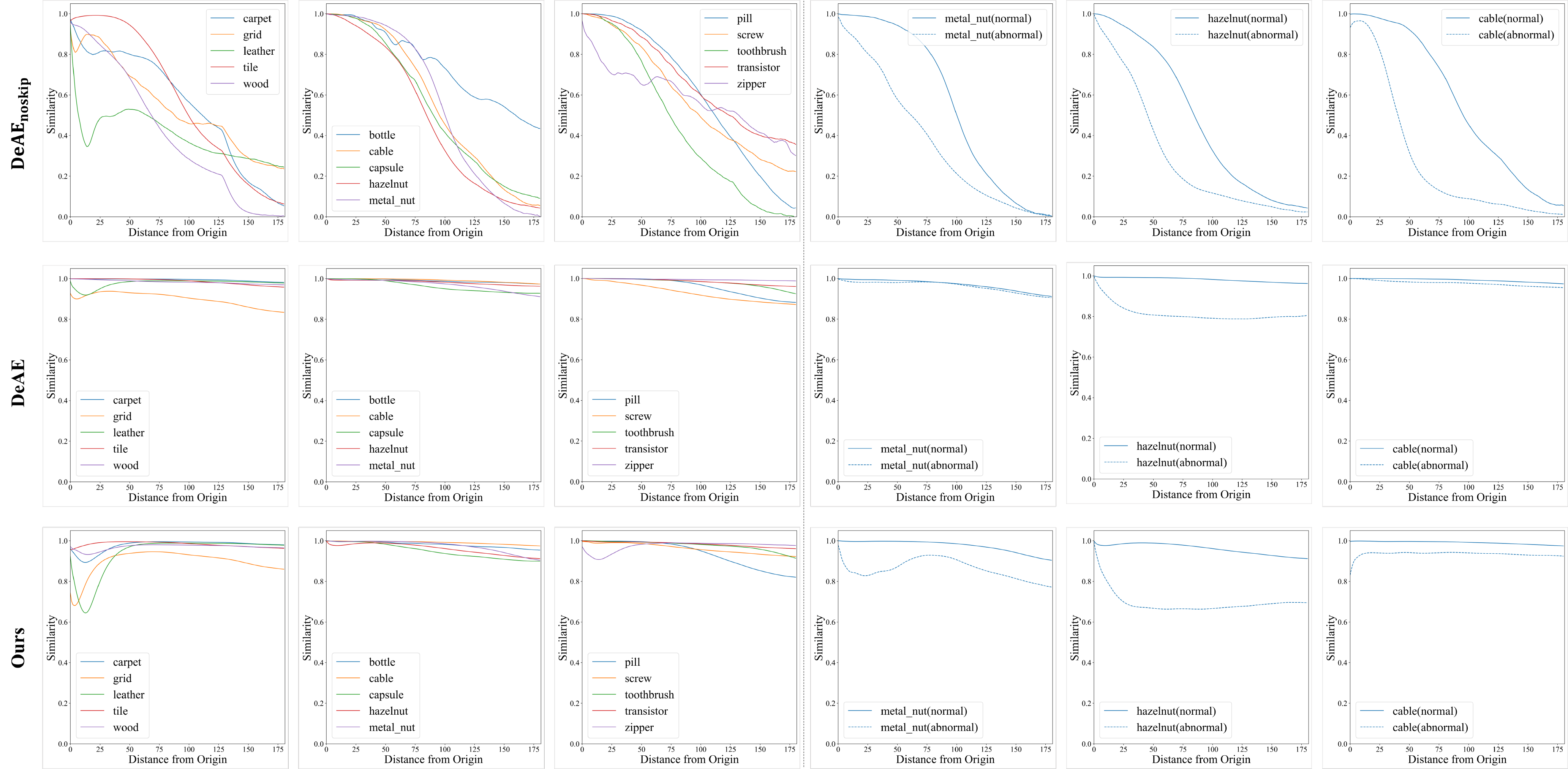}
	\caption{The normal (the left part) and abnormal (the right part) reconstruction error biases in the frequency domain. The increased distance from the origin indicates a higher frequency value. The decreased similarity indicates larger reconstruction errors.}
	\label{frequencybiasplots}
\end{figure*}

\subsection{Results}
Table. \ref{MVTec_det} and Table. \ref{MVTec_pixel} show quantitative comparisons of the anomaly detection and segmentation performance on the MVTec AD dataset. Table. \ref{Visa_image} and Table. \ref{Visa_pixel} show the same comparisons on the VisA dataset. It's observed that our method achieves superior performance for both anomaly detection and segmentation tasks. Furthermore, we conduct a more comprehensive comparison in Table. \ref{OverallComparison}, including the average performance on both the two datasets, the inference speeds, and the model parameters. We choose several current state-of-the-art methods as our baselines, involving both image reconstruction-based methods and feature-based methods. For our method, we progressively reduce the network's base width c from the original 128 to 64 and 32 to observe the trade-off between the performance and efficiency. As shown in Table. \ref{OverallComparison}, overall, our method achieves higher average performance along with higher efficiency. For example, with a very lightweight base network (c=32), we can still outperform the state-of-the-art image reconstruction model DSR by a large margin, i.e., 1.4 \%  AUROC for detection and 12 \% AUPRO for segmentation. Meanwhile, compared to DSR, our method is 2 times faster and uses only 11\% parameters. We also outperform feature-based Patchore and RD4AD both in performance and efficiency while our model is trained from scratch and does not rely on powerful pre-trained models as they do. 

\begin{table}[htb]
\footnotesize
\caption{Quantitative comparison of the average performance and the model efficiency on the MVTec AD and VisA benchmarks. The methods with *  are image reconstruction-based methods.}
\label{OverallComparison}
\centering
\begin{tabular}{ccccc}
\hline
Method & Det & Loc & FPS & Para \\ \hline
PatchCore \cite{roth2022towards}& 97 & (98.5,92.6) & 34 & 82MB \\
RD4AD \cite{deng2022anomaly}& 97.2 & (98.1,93.4) & 37.5 & 89.5MB \\
AST \cite{rudolph2023asymmetric}& 96.0 & (95.2,79.5) & 38 & 147MB \\
DRAEM* \cite{zavrtanik2021draem}& 94.9 & (95.3,85.4) & 45 & 93MB \\
DSR* \cite{zavrtanik2022dsr}& 95 & (93.6,79.9) & 43 & 38.4MB \\
Ours*(c=128) & 97.7 & (98.5,92.7) & 57 & 65.8MB \\
Ours*(c=64) & 97.3 & (98.5,92.7) & 100 & 16.5MB \\
Ours*(c=32) & 96.4 & (98.4,91.9) & 132 & 4.1MB \\ \hline
\end{tabular}
\end{table}

 \section{Discussion and ablation studies.}
\subsection{Frequency biases in image reconstruction}
In the introduction, we argue that normal reconstruction errors generally exhibit a bias toward high-frequency ranges due to the downsampling operation and MSE loss, while abnormal reconstruction errors may encompass all frequency ranges due to anomalies' inherent uncertainty. To verify this, we conduct quantitative experiments where we calculate the cosine similarity between the original and reconstructed images in the frequency domain. We employ three models including DeAE, $\rm DeAE_{noskip}$, and our original restoration model. Among them, DeAE is a conventional denoising reconstruction model that shares the same training paradigm as our original method except for removing our restoration design. $\rm DeAE_{noskip}$ additionally removes the skip connections in our original network. The left part of Fig. \ref{frequencybiasplots} shows the similarity curves of normal test samples in the MVTec AD dataset. It's observed that for vanilla image reconstruction models DeAE and $\rm DeAE_{noskip}$, the normal original and reconstructed images have the highest similarity in low-frequency ranges near the origin. As the frequency increases, the similarity generally begins to decline, demonstrating that normal reconstruction errors are typically biased to high-frequency ranges. The right part of Fig. \ref{frequencybiasplots} shows the abnormal samples' frequency similarity. Although the abnormal similarity also decreases with increasing frequency, it's worth noting that the abnormal samples' frequency similarity differs from that of normal samples in all frequency ranges, demonstrating the uncertainty of abnormal reconstruction errors in the frequency domain. Besides, it's noticed that adopting skip connections can significantly reduce normal high-frequency reconstruction errors. However, for vanilla reconstruction models, this also decreases the abnormal reconstruction distinguishability. For example, DeAE's similarity curves of the normal and abnormal samples do not show significant differences in the metal nut and cable categories. Comparatively, our original restoration model demonstrates superiority including the less normal high-frequency loss while also ensuring the distinguishability of the abnormal reconstruction.     

\subsection{Effectiveness of the frequency restoration task}
We further conduct ablation experiments to evaluate the anomaly detection performance of DeAE and $\rm DeAE_{noskip}$. Meanwhile, since our original restoration task also removes the color information, we introduce another variant termed ColorRe, which does not involve the original frequency restoration but only restores the color information of the grayscale image. The results are shown in Table. \ref{TableDeAE}. It's observed that our frequency restoration task can effectively improve the overall anomaly detection performance.  

\begin{table}[h]
\caption{Quantitative comparison of the anomaly detection performance with different reconstruction tasks. I-AUROC and P-AUROC refer to image-level and pixel-level AUROC respectively.}
\setlength{\tabcolsep}{4.4mm}{
\label{TableDeAE}
\scalebox{0.8}{
\begin{tabular}{c|ccc}
\hline
Method     & I-AUROC\% & P-AUROC\% & AUPRO\% \\ \hline
DeAE       & 95.5      & 96.4      & 89.8    \\
$\rm DeAE_{noskip}$ & 95.5      & 96.4      & 89.8    \\
ColorRe    & 96.4      & 96.8      & 91.3    \\
Ours       & \textbf{98.6}      & \textbf{98.2}      & \textbf{94.0}    \\ \hline
\end{tabular}}}
\end{table}

\subsection{Frequency-domain filter desings}
\begin{figure}[h]
    \centering
		\includegraphics[width=\columnwidth]{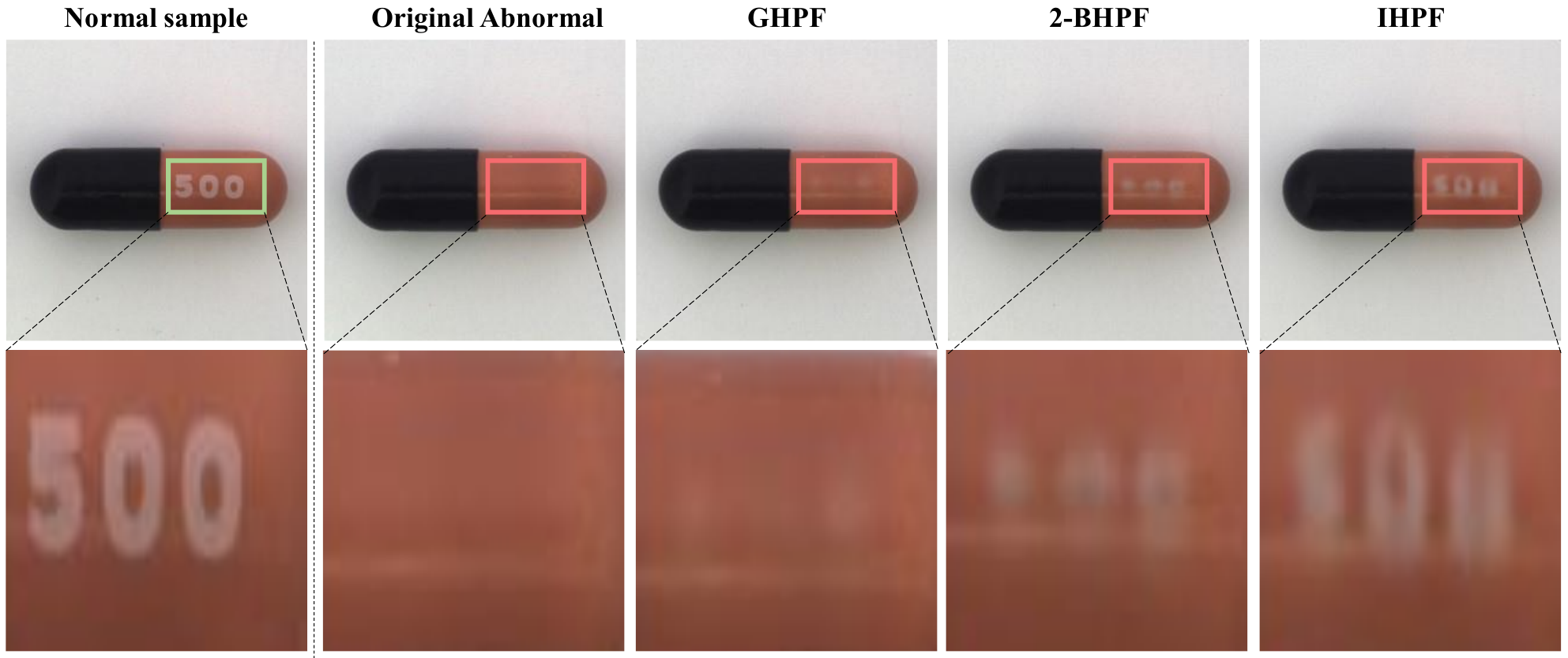}
	\caption{Qualitative comparison of the abnormal reconstruction distinguishability with different frequency domain filters.}
	\label{distguishability3filters}
\end{figure}

\textbf{Filter smoothness.} To evaluate the impacts of the filter smoothness on the final results, we here conduct experiments using IHPF, GHPF, and the third-order BHPF (denoted as 3-BHPF) to replace the original second-order BHPF (2-BHPF) as the high-frequency extractor. The results are shown in Table. \ref{TableFrequencyfilters}. Through comparing the normal reconstruction errors, it's concluded that the smooth filter, e.g., GHPF benefits the normal reconstruction fidelity, as it does not completely remove the low-frequency information but only reduces the amplitude. Another reason may be that the smooth filter generates fewer ringing artifacts which corrupt normal image information. Yet, the anomaly detection task also needs to consider abnormal reconstruction distinguishability. In this regard, the filter with a hard cutoff, e.g., IHPF, is more suitable since it removes more low-frequency information and generates stronger artifacts. These help reduce the identity mapping of abnormal information, as shown in Fig. \ref{distguishability3filters}. In our task, the 2-BHPF achieves the overall optimal performance because it has moderate smoothness which balances the normal reconstruction fidelity and abnormal reconstruction distinguishability. In specific scenarios, it's preferable to select the frequency domain filters based on domain knowledge.

\begin{table}[h]
\caption{Quantitative comparison of the anomaly detection performance with different frequency domain filters. I-AUROC and P-AUROC refer to image-level and pixel-level AUROC respectively. The normal reconstruction errors (Normal errors) are evaluated with MSE in units of 1e-4.}
\setlength{\tabcolsep}{1.6mm}{
\label{TableFrequencyfilters}
\scalebox{0.8}{
\begin{tabular}{c|cccc}
\hline
Extractor & I-AUROC\% & P-AUROC\% & AUPRO\% & Normal errors \\ \hline
IHPF      & 96.0      & 97.2      & 91.3    & 1.50           \\
GHPF      & 98.2      & 97.7      & 93.3    & 0.90          \\
3-BHPF    & 98.3      & 98.2      & 93.9    & 1.15           \\
2-BHPF    & \textbf{98.6}      & \textbf{98.2}      & \textbf{94.0}    & 1.06           \\ \hline
\end{tabular}}}
\end{table}

\begin{figure}[h]
    \centering
		\includegraphics[width=\columnwidth]{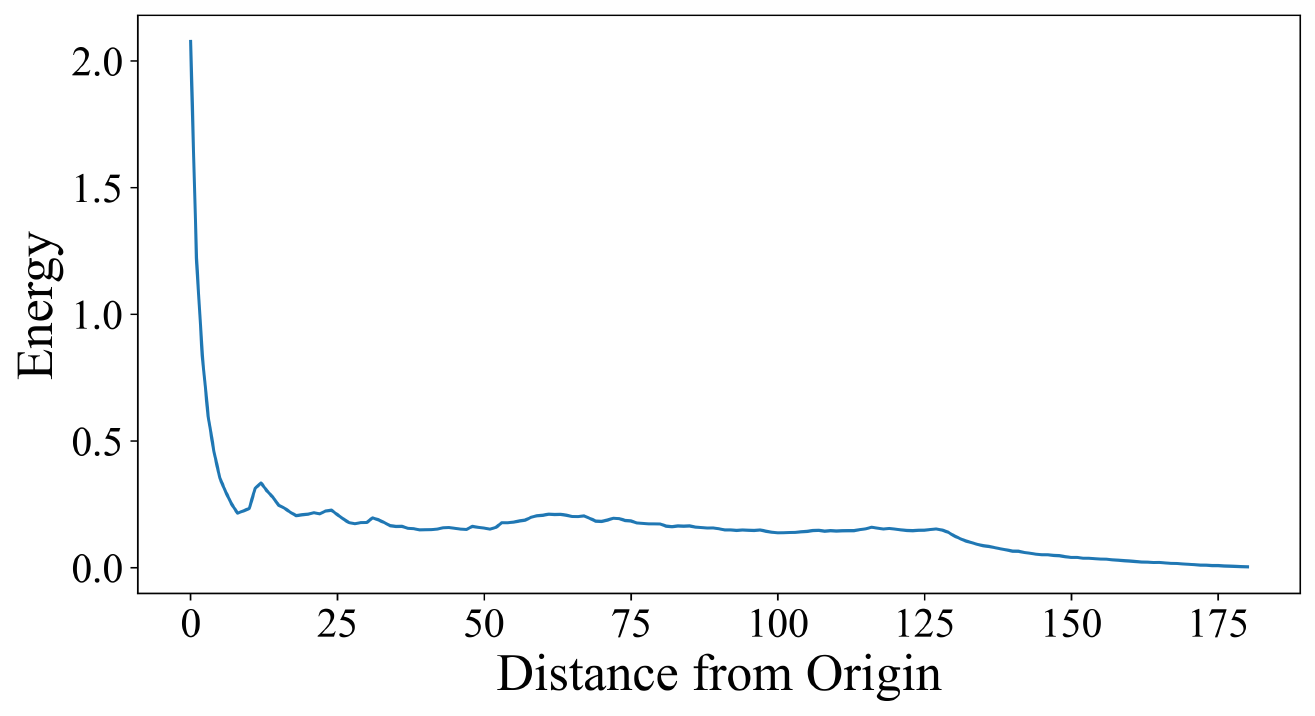}
	\caption{The image energy distribution under different frequency ranges of the MVTec AD dataset. The increased distance from the origin signifies a higher frequency.}
	\label{energy}
\end{figure}

\textbf{Cutoff frequency.} Table. \ref{Cutofffrequencies} shows the comparison of the results using the 2-BHPF with different cutoff frequencies. Similar to the filter smoothness, the cutoff frequency can also affect the reconstruction errors. However, the effect may be less prominent compared to the filter smoothness. For example, the reconstruction error for 2-BHPF with $D_0=10$ is still larger than that for GHPF with $D_0=30$. Also, the reconstruction error for 2-BHPF with $D_0=90$ is still smaller than that for IHBF with $D0=30$. The reason may be that the filter smoothness has a huge impact on the frequency information near the frequency domain origin, which usually takes up most of the energy of the image, as shown in Fig. \ref{energy}.

\begin{table}[h]
\caption{Quantitative comparison of the anomaly detection performance with different cutoff frequencies. I-AUROC and P-AUROC refer to image-level and pixel-level AUROC respectively. The normal reconstruction errors (Normal errors) are evaluated with MSE in units of 1e-4.}
\setlength{\tabcolsep}{2.5mm}{
\label{Cutofffrequencies}
\scalebox{0.8}{
\begin{tabular}{c|cccc}
\hline
$D_0$ & I-AUROC\% & P-AUROC\% & AUPRO\% & Normal errors \\ \hline
10 & 97.6      & 98.0      & 93.1    & \textbf{0.99}          \\
30 & \textbf{98.6}      & 98.2      & \textbf{94.0}    & 1.06          \\
60 & 98.1      & \textbf{98.3}      & \textbf{94.0}    & 1.15          \\
90 & 98,1      & 98.1      & 93.6    & 1.21          \\ \hline
\end{tabular}}}
\end{table}

\textbf{Ringing artifacts.} Another concern is the impact of the ringing artifacts. Due to the inherent coupling of the ringing artifact with frequency characteristics, it may not be possible to analyze their impact in isolation. However, we speculate that the ringing artifact may have less impact on the restoration task compared to the removed frequency information. This is because for 2-BHPF, increasing the cutoff frequency will decrease the ringing artifacts. At this point, if the ringing artifact is dominant, the image restoration task should become simpler, corresponding to lower normal reconstruction errors. Yet, the opposite is observed, indicating that the removed frequency information plays a more crucial role. 

\subsection{Spatial domain high-frequency extractor}
\label{sec5.4}
\begin{figure}[h]
    \centering
		\includegraphics[width=\columnwidth]{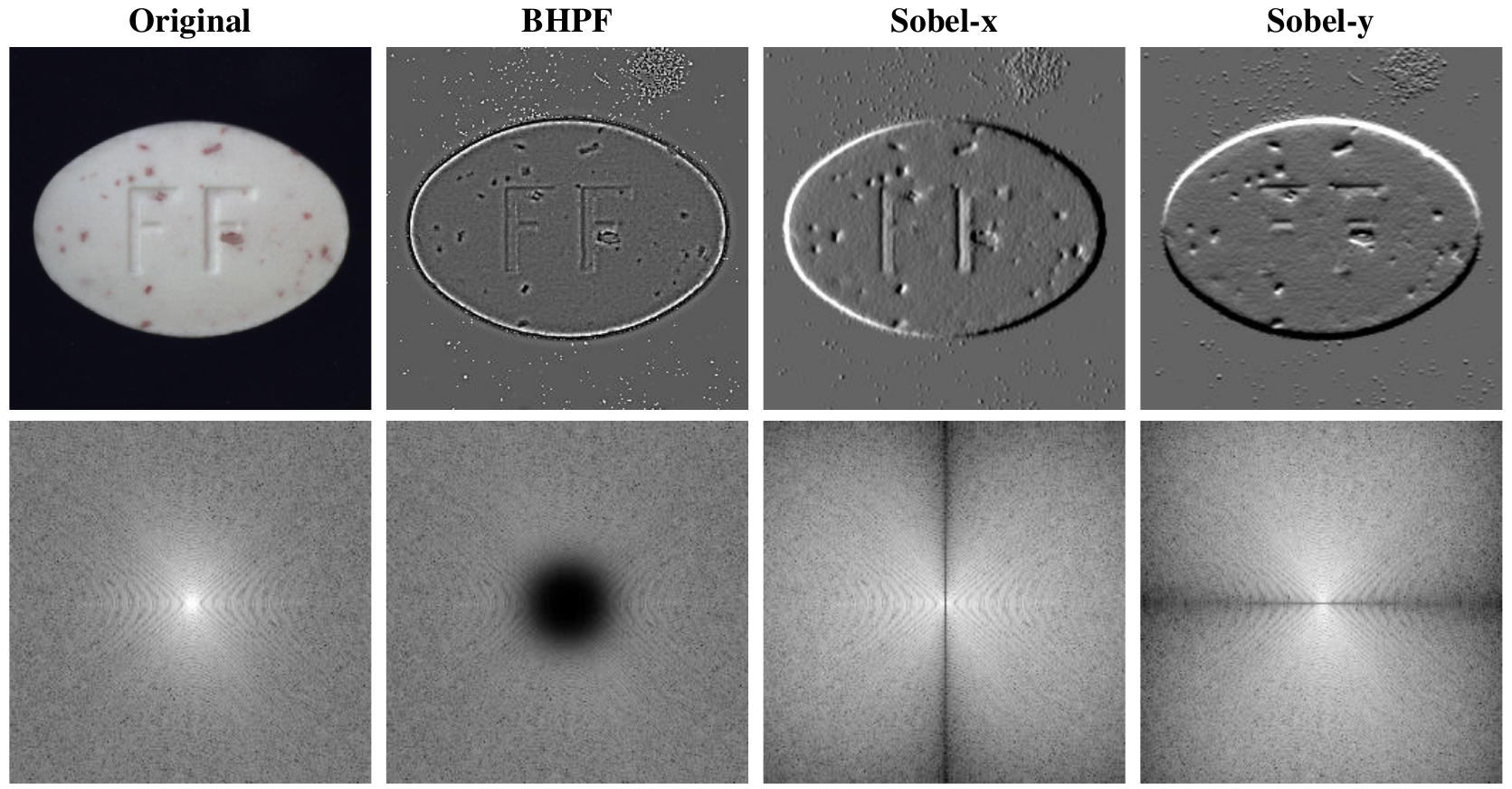}
	\caption{The examples of different spatial ringing artifacts corresponding to different frequency domain filters.}
	\label{BHPFgxgy}
\end{figure}

Apart from frequency-domain filters, it's also feasible to apply the spatial domain image derivatives, in other words, image gradient, to extract high-frequency information. The basic formulation of image gradient can be written as:
\begin{equation}
\bigtriangledown f(x,y) = \begin{bmatrix}g_x
 \\
g_y
\end{bmatrix}=\begin{bmatrix}\frac{\partial f}{\partial x} 
 \\
\frac{\partial f}{\partial y} 
\end{bmatrix}
\label{100}
\end{equation}
where $f(x,y)$ is the image function. $x$ and $y$ are the image spatial coordinates. $g_x =\frac{\partial f}{\partial x}$ and $g_y = \frac{\partial f}{\partial y}$ are the image gradients along different directions. For example, if we apply $g_x$ to obtain the highpass image $f(x,y)_h$, a common formula is:

\begin{equation}
    f(x,y)_h = f(x+1,y)-f(x-1,y)
    \label{gradienthighpass}
\end{equation}
To compare it with the aforementioned frequency-domain filters, we perform 2D DFT $\mathcal{F}[\cdot]$ on eq. \ref{gradienthighpass}, thus obtaining:
\begin{equation}
    \mathcal{F}[f(x,y)_h] = \mathcal{F}[f(x+1,y)-f(x-1,y)]
\end{equation}
Based on the fundamental theorems of the Fourier transform:
\begin{equation}
    \mathcal{F}[ab+cd]=a\mathcal{F}[b]+c\mathcal{F}[d]
\end{equation}
and 
\begin{equation}
    \mathcal{F}[f(x-x_0,y-y_0)]=\mathcal{F}[f(x,y)]e^{-j2\pi(x_0u/M+y_0v/N)}
\end{equation}
, eq. \ref{gradienthighpass} can be transformed into:
\begin{equation}
    F(u,v)_h = F(u,v)[e^{j2\pi u/M}-e^{-j2\pi u/M}]
    \label{gradienttransform}
\end{equation}
Different from the aforementioned frequency-domain filters that only change the amplitude, in eq. \ref{gradienttransform}, it's observed that the gradient operator can change both the amplitude and phase of frequency components. Also, eq. \ref{gradienttransform} shows that the gradient operator exhibits directionality in frequency filtering. For example, eq. \ref{gradienttransform} removes all the frequency information along the axis $u=0$. Fig. \ref{BHPFgxgy} provides a qualitative comparison of such directionality when using $g_x$, $g_y$, and BHPF respectively. In this regard, when utilizing image gradients to formulate image restoration tasks, we can adjust the input gradient directions to control the preservation of the frequency information. Besides, for the gradient operator, we can also adjust the kernel size. A large kernel size has a mean filtering effect which will remove more high-frequency information. 

Experimentally, we find that using the concatenation of $g_x$ and $g_y$ obtained from the $3 \times 3$ Sobel operators as input can achieve the best performance on the selected benchmarks, as shown in Table. \ref{FAIRGradientxy3} (denoted as ${\rm FAIR}_{xy3}$).  Table. \ref{FAIRGradienvariants} shows the performance of its variants including ${\rm FAIR}_{x3}$, which only uses the $g_x$ as input and ${\rm FAIR}_{xy5}$, which modifies the kernel size to 5.

\begin{table}[h]
\caption{Quantitative evaluation of the anomaly detection performance using the Sobel gradient operator as a high-frequency extractor. I-AUROC and P-AUROC refer to image-level and pixel-level AUROC respectively.}
\setlength{\tabcolsep}{4.0mm}{
\label{FAIRGradientxy3}
\scalebox{0.8}{
\begin{tabular}{c|ccc}
\hline
\multirow{2}{*}{Dataset} & \multicolumn{3}{c}{${\rm FAIR}_{xy3}$}     \\ \cline{2-4} 
                         & I-AUROC\% & P-AUROC\% & AUPRO\% \\ \hline
MVTec AD                 & 98.5      & 97.2      & 92.8    \\
VisA                     & 96.6      & 98.5      & 91.0    \\
Average                  & 97.6      & 97.9      & 91.9    \\ \hline
\end{tabular}}}
\end{table}

\begin{table}[h]
\caption{Quantitative comparison of the anomaly detection performance with different variants of the Sobel gradient operators. I-AUROC and P-AUROC refer to image-level and pixel-level AUROC respectively. The normal reconstruction errors (Normal errors) are evaluated with MSE in units of 1e-4.}
\setlength{\tabcolsep}{1.5mm}{
\label{FAIRGradienvariants}
\scalebox{0.8}{
\begin{tabular}{c|cccc}
\hline
Method  & I-AUROC\% & P-AUROC\% & AUPRO\% & Normal errors \\ \hline
${\rm FAIR}_{x3}$  & 98.2      & \textbf{97.9 }     & \textbf{93.8}    & 1.00          \\
${\rm FAIR}_{xy5}$ & 98.0      & 97.3      & 93.0    & 0.93          \\
${\rm FAIR}_{xy3}$ & \textbf{98.5}      & 97.2      & 92.8    & \textbf{0.90}         \\ \hline
\end{tabular}}}
\end{table}

\subsection{Anomaly evaluation function}
Here, we explore the effectiveness of different anomaly evaluation functions including MSGMS and the proposed color difference (Co). Additionally, we also attempt to combine our restoration network with a discriminative network (Dis) following the same training paradigm in \cite{zavrtanik2021draem}. As shown in Table. \ref{evaluationfunction}, the proposed color difference can improve the performance of MSGMS but cannot be used alone since many anomalies do not manifest as color differences. The discriminative network here is not superior to hand-crafted explicit functions.

\begin{table}[h]
\caption{Quantitative comparison of the anomaly detection performance of our method with different anomaly evaluation functions. I-AUROC and P-AUROC refer to image-level and pixel-level AUROC respectively.}
\setlength{\tabcolsep}{3.8mm}{
\label{evaluationfunction}
\scalebox{0.8}{
\begin{tabular}{c|ccc}
\hline
Function & I-AUROC\% & P-AUROC\% & AUPRO\% \\ \hline
MSGMS    & 98.4      & 98.0      & 93.6    \\
Co       & 87.2      & 92.5      & 79.3    \\
Dis      & 98.2      & 97.2      & 89.6    \\
MSGMS+Co & \textbf{98.6}      & \textbf{98.2}      & \textbf{94.0}    \\ \hline
\end{tabular}}}
\end{table}

\begin{table}[h]
\caption{Quantitative comparison of the anomaly detection performance of our method without using any extra data. I-AUROC and P-AUROC refer to image-level and pixel-level AUROC respectively.}
\setlength{\tabcolsep}{4.0mm}{
\label{Extradata}
\scalebox{0.8}{
\begin{tabular}{c|ccc}
\hline
\multirow{2}{*}{Dataset} & \multicolumn{3}{c}{${\rm FAIR}_{\rm noDTD}$}   \\ \cline{2-4} 
                         & I-AUROC\% & P-AUROC\% & AUPRO\% \\ \hline
MVTec AD                 & 98.1      & 98.2      & 93.6    \\
VisA                     & 97.1      & 98.7      & 91.2    \\
Average                  & 97.9      & 98.5      & 92.4    \\ \hline
\end{tabular}}}
\end{table}

\subsection{Reliance on extra data}
Generally, introducing extra data, such as ImageNet pre-trained weights or off-task datasets, has been demonstrated effective in improving anomaly detection performance. However, it's still popular to explore the model that is trained from scratch and does not require extra data. To this end, we evaluate the performance of a variant denoted as ${\rm FAIR}_{\rm noDTD}$ which replaces the DTD textures with the internal texture of each training image in Sec. \ref{sec3.1}. As shown in Table. \ref{Extradata}, without the aid of any extra data, our method still outperforms existing state-of-the-art methods (see Table. \ref{OverallComparison}) which apply ImageNet pre-trained weights or extra texture datasets. 

\section{Conclusion}
This paper proposes a novel self-supervised frequency image restoration task for industrial visual anomaly detection. The restoration task leverages the different frequency-domain biases between normal and abnormal reconstruction errors. Our method achieves state-of-the-art performance in various industrial scenarios. Meanwhile, it's very simple, using only a vanilla UNet without complex architectural designs. We hope that this simple and effective model can benefit real industrial deployments and that our explorations in the frequency domain perspective can inspire future research.

\section*{Declaration of Competing Interest}
The authors declare that they have no known competing financial interests or personal relationships that could have appeared to influence the work reported in this paper.

\section*{Acknowledgement}
The calculations were performed by using the HPC Platform at Xi’an Jiaotong University. This work is supported by the Open Research Fund (KLMVI-2023-HIT-19) of Anhui Province Key Laboratory of Machine Vision Inspection.

\end{document}